Study of dynamical system based obstacle avoidance via manipulating orthogonal coordinates

Weiya Ren[1,2]

[1] Artificial Intelligence Research Center of National Innovation Institute of Defense Technology
[2] Tianjin Artificial Intelligence Innovation Center, Tianjin, P.R China.
E-mail: weiyren.phd@gmail.com

**Abstract** In this paper, we consider the general problem of obstacle avoidance based on dynamical system. The modulation matrix is developed by introducing orthogonal coordinates, which makes the modulation matrix more reasonable. The new trajectory's direction can be represented by the linear combination of orthogonal coordinates. A orthogonal coordinates manipulating approach is proposed by introducing rotating matrix to solve the local minimal problem and provide more reasonable motions in 3-D or higher dimension space. The proposed method also provide a solution for patrolling around a convex shape. Experimental results on several designed dynamical systems demonstrate the effectiveness of the proposed approach.

## I.   INTRODUCTION

Obstacle avoidance is widely used in many research fields, including path planning [2], tracking [3], controlling [4], and others [8][9]. Recently, some powerful approaches to obstacle avoidance have been proposed such as deep reinforcement learning [10], potential field method [11], harmonic potential function [12], etc. Different from those methods, dynamical system based approach offer an effective approach for obstacle avoidance [1][5], which can help to react to the unknown obstacles and external perturbations in real time. Furthermore, the obstacles are allowed to be represented by either analytical or point cloud [16]. Dynamical system based approach does not need re-plan process and it can guarantee the obstacles impenetrability. The Dynamical system is always provided by the user, which can be ground robot DS [15], and UVA DS [14] which can also be defined by vector field methods such as LGVF [6], CLGVF [13], TGVF[7].

[1] proposed a novel method by introducing a modulation matrix, which is developed in this paper by a novel approach named Obstacle Avoidance via Manipulating Orthogonal Coordinates(OA-MOC). In contrast to [1], our contribution is twofold. First, an orthogonal coordinates are introduced to construct the modulation matrix. The orthogonal coordinates can be seen as a orthogonal basis matrix, which can be used to represent the modulated system. The new direction of the trajectory motion is only decided by the angles between the original trajectory's direction and the orthogonal basis. Second, a orthogonal coordinates manipulating approach is proposed by rotating basis, which can solve the local minimal problem and provide more reasonable motions in 3-D or higher dimension space. The proposed method also provide a solution for patrolling around a convex shape.

The rest of the paper is organized as follows: Section 2 gives a brief review of related concepts. In Section 3, we study the modulation matrix by introducing the orthogonal basis matrix, also named orthogonal coordinates. Section 4 propose a manipulating approach to the orthogonal coordinates. Experimental results are presented in Section 5. Finally, conclusions are drawn in Section 6.

## II. RELATED CONCEPTS

$\xi \in \mathbb{R}_d$ is the state variable whose temporal evolution may be governed by time-invariant or time-varying DS by:

$$\dot{\xi} = f(\xi), \quad f: \mathbb{R}_d \mapsto \mathbb{R}_d. \qquad \text{autonomous DS} \quad (1)$$

$$\dot{\xi} = f(t, \xi), \quad f: \mathbb{R}^+ \times \mathbb{R}_d \mapsto \mathbb{R}_d. \qquad \text{non-auto. DS} \quad (2)$$

A continuous function $\Gamma(\xi)$ is defined in [1] that projects $\mathbb{R}_d$ to $\mathbb{R}$, which can distinguish three regions which are points inside the obstacle, at its boundary, and outside the obstacle, respectively. $\Gamma(\xi)$ is defined by:

$$\Gamma(\xi) = \sum_{i=1}^{d} (\xi_i - \xi_i^{center})^{2p_i} / a_i. \qquad (3)$$

where $p_i, a_i, i = 1,2..,d$ are obstacles' parameters. $\Gamma(\xi) = 1$ when point $\xi$ is at its boundary, $\Gamma(\xi) < 1$ when point $\xi$ is inside the obstacle, and $\Gamma(\xi) > 1$ when point $\xi$ is outside the obstacle.

By introducing a modulation matrix in [1], the original DS is modified by:

$$\dot{\xi} = M^{d \times d}(\xi) f(\xi). \qquad (4)$$

where $M^{d \times d} \in R^{d \times d}$. $\dot{\xi}, f(\xi) \in R^{d \times 1}$, $\dot{\xi}$ is the first time derivative of $\xi$. $f(\xi)$ is an autonomous DS, which also can be defined by the non-auto. DS.

The desire trajectory can be calculated by integrating $\dot{\xi}$:

$$\xi_{t+1} = \xi_t + \dot{\xi} \delta_t. \qquad (5)$$

where $\delta_t$ is the integration time step.

## III. ORTHOGONAL BASIS MATRIX

### A. Coordinates representation

We first consider 2-D situation, i.e. $d = 2$. In [1], a modulation matrix is defined as:

$$M^{2 \times 2}(\xi) = \hat{E}^{2 \times 2}(\xi) D^{2 \times 2}(\xi) \hat{E}^{2 \times 2}(\xi)^{(-1)} \qquad (6)$$

with the matrix of basis vectors $\hat{E}^{2 \times 2}(\xi)$ and the associated eigenvalues $D^{2 \times 2}(\xi)$:

$$\hat{E}^{2 \times 2}(\xi) = [e_1(\xi), e_2(\xi)]. \qquad (7)$$

where $e_1(\xi) = \left[\frac{\partial \Gamma(\xi)}{\xi_1}, \frac{\partial \Gamma(\xi)}{\xi_2}\right]^T$ is the normal vector and $e_2(\xi) = \left[\frac{\partial \Gamma(\xi)}{\xi_2}, -\frac{\partial \Gamma(\xi)}{\xi_1}\right]^T$ is a vertical vector of $e_1(\xi)$.

$$D^{2 \times 2}(\xi) = \begin{bmatrix} \lambda_1(\xi) & 0 \\ 0 & \lambda_2(\xi) \end{bmatrix}. \qquad (8)$$

where

$$\begin{cases} \lambda_1(\xi) = 1 - \dfrac{1}{|\Gamma(\xi)|} \\ \lambda_2(\xi) = 1 + \dfrac{1}{|\Gamma(\xi)|} \end{cases}. \qquad (9)$$

If $|\Gamma(\xi)| \to +\infty$, then $D(\xi) \to I$, and $M(\xi) = \hat{E}^{2 \times 2}(\xi) D(\xi) \hat{E}^{2 \times 2}(\xi)^{(-1)} \to I$. Note that $\hat{E}^{2 \times 2}(\xi)^{(-1)}$ requires $|\hat{E}^{2 \times 2}(\xi)| \neq 0$, which means $|\hat{E}^{2 \times 2}(\xi)| = -\left[\left(\frac{\partial \Gamma(\xi)}{\xi_1}\right)^2 + \left(\frac{\partial \Gamma(\xi)}{\xi_2}\right)^2\right] \neq 0$.

We denote $\hat{E}^{2\times 2}(\xi)^{(-1)} = [u_1(\xi); u_2(\xi)]$, where $u_1(\xi), u_2(\xi) \in R^{1\times 2}$. Then we have:

$$M^{2\times 2}(\xi) = [e_1(\xi), e_2(\xi)] \begin{bmatrix} 1 - \frac{1}{|\Gamma(\xi)|} & 0 \\ 0 & 1 + \frac{1}{|\Gamma(\xi)|} \end{bmatrix} [u_1(\xi); u_2(\xi)]$$

$$= \left(1 - \frac{1}{|\Gamma(\xi)|}\right) e_1(\xi) u_1(\xi) + \left(1 + \frac{1}{|\Gamma(\xi)|}\right) e_2(\xi) u_2(\xi). \tag{10}$$

Since

$$\dot{\xi} = M^{2\times 2}(\xi) f(\xi) = \left(1 - \frac{1}{|\Gamma(\xi)|}\right) e_1(\xi) u_1(\xi) f(\xi) + \left(1 + \frac{1}{|\Gamma(\xi)|}\right) e_2(\xi) u_2(\xi) f(\xi). \tag{11}$$

where $u_1(\xi), u_2(\xi) \in R^{1\times 2}, f(\xi) \in R^{2\times 1}$, $\left(1 - \frac{1}{|\Gamma(\xi)|}\right) u_1(\xi) f(\xi), \left(1 + \frac{1}{|\Gamma(\xi)|}\right) u_2(\xi) f(\xi) \in R$.

Thus

$$\dot{\xi} = \left(1 - \frac{1}{|\Gamma(\xi)|}\right) u_1(\xi) f(\xi) e_1(\xi) + \left(1 + \frac{1}{|\Gamma(\xi)|}\right) u_2(\xi) f(\xi) e_2(\xi). \tag{12}$$

So $\dot{\xi}$ can be expressed as a linear combination of two linear independent vectors $e_1(\xi)$ and $e_2(\xi)$, as seen in Fig. 1.

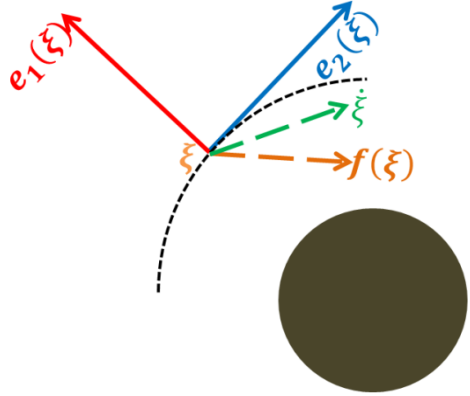

Fig. 1. $\dot{\xi}$ is generated from $f(\xi)$ based on the coordinates $e_1(\xi)$ and $e_2(\xi)$.

B. *Local minimal problem*

Since

$$\hat{E}^{2\times 2}(\xi) = \begin{bmatrix} \frac{\partial \Gamma(\xi)}{\xi_1} & \frac{\partial \Gamma(\xi)}{\xi_2} \\ \frac{\partial \Gamma(\xi)}{\xi_2} & -\frac{\partial \Gamma(\xi)}{\xi_1} \end{bmatrix} \tag{13}$$

We have

$$\hat{E}^{2\times2}(\xi)^{(-1)} = [u_1(\xi); u_2(\xi)] = \frac{1}{\left(\frac{\partial \Gamma(\xi)}{\xi_1}\right)^2 + \left(\frac{\partial \Gamma(\xi)}{\xi_2}\right)^2} \begin{bmatrix} \frac{\partial \Gamma(\xi)}{\xi_1} & \frac{\partial \Gamma(\xi)}{\xi_2} \\ \frac{\partial \Gamma(\xi)}{\xi_2} & -\frac{\partial \Gamma(\xi)}{\xi_1} \end{bmatrix}$$

$$= \frac{1}{\left(\frac{\partial \Gamma(\xi)}{\xi_1}\right)^2 + \left(\frac{\partial \Gamma(\xi)}{\xi_2}\right)^2} \hat{E}^{2\times2}(\xi)^T$$

$$= \frac{1}{\left(\frac{\partial \Gamma(\xi)}{\xi_1}\right)^2 + \left(\frac{\partial \Gamma(\xi)}{\xi_2}\right)^2} [e_1(\xi)^T; e_2(\xi)^T] \quad (14)$$

From Eq.(12) and Eq.(14), we have

$$\dot{\xi} = \frac{\left(1 - \frac{1}{|\Gamma(\xi)|}\right) f(\xi)^T e_1(\xi)}{\left(\frac{\partial \Gamma(\xi)}{\xi_1}\right)^2 + \left(\frac{\partial \Gamma(\xi)}{\xi_2}\right)^2} e_1(\xi) + \frac{\left(1 + \frac{1}{|\Gamma(\xi)|}\right) f(\xi)^T e_2(\xi)}{\left(\frac{\partial \Gamma(\xi)}{\xi_1}\right)^2 + \left(\frac{\partial \Gamma(\xi)}{\xi_2}\right)^2} e_2(\xi)$$

$$= \frac{\left(1 - \frac{1}{|\Gamma(\xi)|}\right) \|f(\xi)\| \|e_1(\xi)\| \cos(\theta_{f(\xi),e_1(\xi)})}{\left(\frac{\partial \Gamma(\xi)}{\xi_1}\right)^2 + \left(\frac{\partial \Gamma(\xi)}{\xi_2}\right)^2} e_1(\xi)$$

$$+ \frac{\left(1 + \frac{1}{|\Gamma(\xi)|}\right) \|f(\xi)\| \|e_2(\xi)\| \cos(\theta_{f(\xi),e_2(\xi)})}{\left(\frac{\partial \Gamma(\xi)}{\xi_1}\right)^2 + \left(\frac{\partial \Gamma(\xi)}{\xi_2}\right)^2} e_2(\xi). \quad (15)$$

where $\theta_{f(\xi),e_1(\xi)}$ and $\theta_{f(\xi),e_2(\xi)}$ are the angle between $f(\xi)$ and $e_1(\xi)$, $e_2(\xi)$, respectively.

Once we have the basis vectors $e_1(\xi), e_2(\xi)$ of the obstacle, $\dot{\xi}$'s direction is only decided by $\theta_{f(\xi),e_1(\xi)}$ and $\theta_{f(\xi),e_2(\xi)}$. If $f(\xi)$ and $e_2(\xi)$ are vertical and $f(\xi)$ and $e_1(\xi)$ are collinear with converse direction, we have:

$$\dot{\xi} = \frac{-\left(1 - \frac{1}{|\Gamma(\xi)|}\right) \|f(\xi)\| \|e_1(\xi)\|}{\left(\frac{\partial \Gamma(\xi)}{\xi_1}\right)^2 + \left(\frac{\partial \Gamma(\xi)}{\xi_2}\right)^2} e_1(\xi). \quad (16)$$

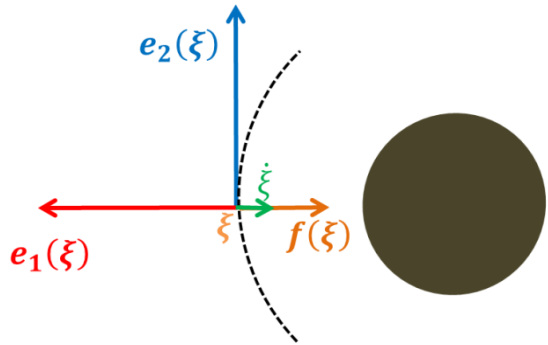

Fig. 2. The local minimal problem.

When the motion $(\{\xi_t\}, t = 0 \dots \infty)$ moves towards to the surface of the obstacle ($|\Gamma(\xi)| \to 1$),

$\dot{\xi}$ becomes to zero ($\dot{\xi} \to 0$) and the local minimal problem happens, which means the motion will stability stay on the surface on the obstacle, as seen in Fig. 2.

C. *Orthogonal Basis matrix*

For 2-D situation, Eq(15) is established only when the modulation matrix is a orthogonal matrix. In the rest of the paper, we adopt the orthogonal basis matrix to construct the modulation matrix.

Due to:

$$M^{2\times 2}(\xi) = \hat{E}^{2\times 2}(\xi)D^{2\times 2}(\xi)\hat{E}^{2\times 2}(\xi)^{(-1)} = \frac{\hat{E}^{2\times 2}(\xi)D(\xi)\hat{E}^{2\times 2}(\xi)^T}{\left(\frac{\partial \Gamma(\xi)}{\xi_1}\right)^2 + \left(\frac{\partial \Gamma(\xi)}{\xi_2}\right)^2}. \quad (17)$$

The orthogonal basis matrix for 2-D situation is defined as follows:

$$E^{2\times 2}(\xi) \triangleq \frac{1}{\sqrt{\left(\frac{\partial \Gamma(\xi)}{\xi_1}\right)^2 + \left(\frac{\partial \Gamma(\xi)}{\xi_2}\right)^2}} \hat{E}^{2\times 2}(\xi). \quad (18)$$

Then we have

$$M^{2\times 2}(\xi) = E^{2\times 2}(\xi)D(\xi)E^{2\times 2}(\xi)^T = E^{2\times 2}(\xi)D(\xi)E^{2\times 2}(\xi)^{(-1)}. \quad (19)$$

In the rest of the paper, the orthogonal basis matrix in 2-D situation is denoted as $E^{2\times 2}(\xi) \triangleq [e_1(\xi), e_2(\xi)]$ for simplicity, and we have $e_i(\xi)^T e_i(\xi) = 1, i = 1,2$.

G. *Multi obstacles*

For 2-D situation, suppose we have $N$ multiple obstacles. The $j$-th obstacle's modulation matrix and basis matrix are denoted as $M^{2\times 2}_{(j)}(\xi)$ and $E^{2\times 2}_{(j)}(\xi)$ ($j = 1,2,\ldots N$), respectively. $E^{2\times 2}_{(j)}(\xi), j = 1,2,\ldots N$ are orthogonal matrix and we denote $E^{2\times 2}_{(j)}(\xi) = [e_{(j)1}(\xi), e_{(j)2}(\xi)], i = 1,2$.

$M^{2\times 2}(\xi)$ can be defined by two different way as follows:

$$M^{2\times 2}(\xi) = \prod_{j=1}^{N} M^{2\times 2}_{(j)}(\xi). \quad (20)$$

where

$$D^{2\times 2}_{(j)}(\xi) = \begin{bmatrix} 1 - \frac{w_j}{|\Gamma(\xi)|} & 0 \\ 0 & 1 + \frac{w_j}{|\Gamma(\xi)|} \end{bmatrix}. \quad (21)$$

Or

$$M^{2\times 2}(\xi) = \sum_{j=1}^{N} w_j M^{2\times 2}_{(j)}(\xi). \quad (22)$$

where

$$D^{2\times 2}_{(j)}(\xi) = \begin{bmatrix} 1 - \frac{1}{|\Gamma(\xi)|} & 0 \\ 0 & 1 + \frac{1}{|\Gamma(\xi)|} \end{bmatrix}. \quad (23)$$

Note that $w_j, j = 1,2,\ldots,N$ in E.q(21) and E.q(22) are the weighting coefficients according

to [1]:

$$w_j = \prod_{i=1,i\neq j}^{N} \frac{\Gamma_i(\xi) - 1}{\Gamma_j(\xi) - 1 + \Gamma_i(\xi) - 1}, \quad j = 1,2,\ldots,N \quad (24)$$

where $\Gamma_j(\xi), j = 1,2,\ldots,N$ is the $j$-th obstacle's continuous function. $0 \leq w_j \leq 1$, $j = 1,2,\ldots,N$. We further normalize $w_j$ to make $\sum_{j=1}^{N} w_j = 1$.

Suppose we have two obstacles, i.e. $N = 2$, and define $M^{2\times 2}(\xi)$ according to Eq.(20), then we have:

$$M^{2\times 2}(\xi) = \prod_{j=1}^{2} M_{(j)}^{2\times 2}(\xi)$$

$$= \left(1 - \frac{w_1}{|\Gamma(\xi)|}\right)\left(1 - \frac{w_2}{|\Gamma(\xi)|}\right) \cos\left(\theta_{e_{(1)1}(\xi),e_{(2)1}(\xi)}\right) e_{(1)1}(\xi) e_{(2)1}(\xi)^T$$

$$+ \left(1 - \frac{w_1}{|\Gamma(\xi)|}\right)\left(1 + \frac{w_2}{|\Gamma(\xi)|}\right) \cos\left(\theta_{e_{(1)1}(\xi),e_{(2)2}(\xi)}\right) e_{(1)1}(\xi) e_{(2)2}(\xi)^T$$

$$+ \left(1 + \frac{w_1}{|\Gamma(\xi)|}\right)\left(1 - \frac{w_2}{|\Gamma(\xi)|}\right) \cos\left(\theta_{e_{(1)2}(\xi),e_{(2)1}(\xi)}\right) e_{(1)2}(\xi) e_{(2)1}(\xi)^T$$

$$+ \left(1 + \frac{w_1}{|\Gamma(\xi)|}\right)\left(1 + \frac{w_2}{|\Gamma(\xi)|}\right) \cos\left(\theta_{e_{(1)2}(\xi),e_{(2)2}(\xi)}\right) e_{(1)2}(\xi) e_{(2)2}(\xi)^T ..(25)$$

So $\dot\xi$ can be computed by:

$$\dot\xi = \prod_{j=1}^{2} M_{(j)}^{2\times 2}(\xi) f(\xi)$$

$$= \left(1 - \frac{w_1}{|\Gamma(\xi)|}\right) \|f(\xi)\| \left\{\left(1 - \frac{w_2}{|\Gamma(\xi)|}\right) \cos\left(\theta_{e_{(1)1}(\xi),e_{(2)1}(\xi)}\right) \cos\left(\theta_{f(\xi),e_{(2)1}(\xi)}\right)\right.$$

$$+ \left(1 + \frac{w_2}{|\Gamma(\xi)|}\right) \cos\left(\theta_{e_{(1)1}(\xi),e_{(2)2}(\xi)}\right) \cos\left(\theta_{f(\xi),e_{(2)2}(\xi)}\right)\right\} e_{(1)1}(\xi)$$

$$+ \left(1 + \frac{w_1}{|\Gamma(\xi)|}\right) \|f(\xi)\| \left\{\left(1 - \frac{w_2}{|\Gamma(\xi)|}\right) \cos\left(\theta_{e_{(1)2}(\xi),e_{(2)1}(\xi)}\right) \cos\left(\theta_{f(\xi),e_{(2)1}(\xi)}\right)\right.$$

$$+ \left(1 + \frac{w_2}{|\Gamma(\xi)|}\right) \cos\left(\theta_{e_{(1)2}(\xi),e_{(2)2}(\xi)}\right) \cos\left(\theta_{f(\xi),e_{(2)2}(\xi)}\right)\right\} e_{(1)2}(\xi). \quad (26)$$

If $M^{2\times 2}(\xi)$ is defined by Eq.(22), we have:

$$\dot\xi = M^{2\times 2}(\xi) f(\xi) = \sum_{j=1}^{N} w_j M_{(j)}^{2\times 2}(\xi) f(\xi)$$

$$= \left(1 - \frac{1}{|\Gamma(\xi)|}\right) \|f(\xi)\| \left[w_1 \cos\left(\theta_{f(\xi),e_{(1)1}(\xi)}\right) e_{(1)1}(\xi) + w_2 \cos\left(\theta_{f(\xi),e_{(2)1}(\xi)}\right) e_{(2)1}(\xi)\right]$$

$$+ \left(1 + \frac{1}{|\Gamma(\xi)|}\right) \|f(\xi)\| \left[w_1 \cos\left(\theta_{f(\xi),e_{(1)2}(\xi)}\right) e_{(1)2}(\xi) + w_2 \cos\left(\theta_{f(\xi),e_{(2)2}(\xi)}\right) e_{(2)2}(\xi)\right]$$

$$\triangleq \left(1 - \frac{1}{|\Gamma(\xi)|}\right) \|f(\xi)\| \cos(\theta_{f(\xi),e_1(\xi)}) e_1(\xi) + \left(1 + \frac{1}{|\Gamma(\xi)|}\right) \|f(\xi)\| \cos(\theta_{f(\xi),e_2(\xi)}) e_2(\xi). (27)$$

**Proposition 1** $e_1(\xi), e_2(\xi)$ defined by Eq.(27) are perpendicular, and satisfies $e_i(\xi)^T e_i(\xi) = 1, i = 1,2$.

Form Eq.(26) and Eq.(27) we can observe that, $\dot{\xi}$ can be expressed as a linear combination of two unit perpendicular vectors. $\dot{\xi}$'s direction is decided by angles among obstacles' basis vectors and angles between $f(\xi)$ and obstacles' basis vectors.

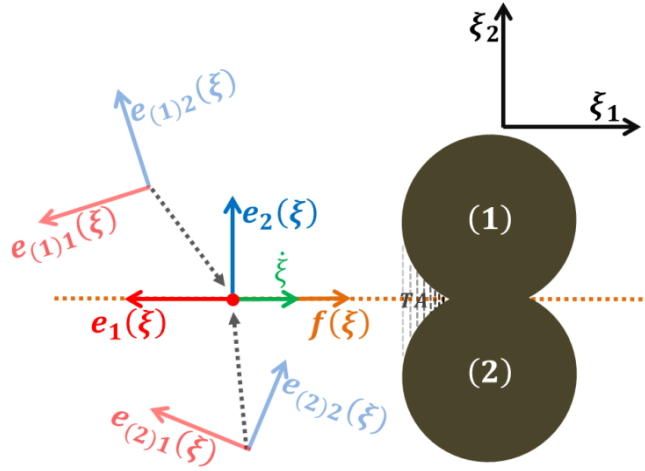

Fig. 3. Multi obstacles case and the trap area(TA) appears.

Supposed we have two horizontal mirror obstacles Obs.(1) and Obs.(2), as seen in Fig. 3. Then the compound basis vectors $e_1(\xi), e_2(\xi)$ makes the trajectory move towards to the trap area. In fact, there are many ways for the trajectory enter into the trap area(TA), which depends on $f(\xi)$ and the compound basis vectors.

*H. 3-D or higher dimensions*

We prefer to select basis vectors of $E$ that satisfies Eq.(19), which means $E$ should be a orthogonal matrix that $E^T = E^{(-1)}$. In 2-D case,

$$E^{2\times 2}(\xi) = \frac{1}{\sqrt{\left(\frac{\partial \Gamma(\xi)}{\xi_1}\right)^2 + \left(\frac{\partial \Gamma(\xi)}{\xi_2}\right)^2}} \begin{bmatrix} \frac{\partial \Gamma(\xi)}{\xi_1} & \frac{\partial \Gamma(\xi)}{\xi_2} \\ \frac{\partial \Gamma(\xi)}{\xi_2} & -\frac{\partial \Gamma(\xi)}{\xi_1} \end{bmatrix} \quad (28)$$

For 3-D case, we first construct a orthogonal matrix by:

$$\tilde{E}^{3\times 3}(\xi) = \begin{bmatrix} \frac{\partial \Gamma(\xi)}{\xi_1}/\sqrt{\left(\frac{\partial \Gamma(\xi)}{\xi_1}\right)^2 + \left(\frac{\partial \Gamma(\xi)}{\xi_2}\right)^2} & \frac{\partial \Gamma(\xi)}{\xi_2}/\sqrt{\left(\frac{\partial \Gamma(\xi)}{\xi_1}\right)^2 + \left(\frac{\partial \Gamma(\xi)}{\xi_2}\right)^2} & 0 \\ \frac{\partial \Gamma(\xi)}{\xi_2}/\sqrt{\left(\frac{\partial \Gamma(\xi)}{\xi_1}\right)^2 + \left(\frac{\partial \Gamma(\xi)}{\xi_2}\right)^2} & -\frac{\partial \Gamma(\xi)}{\xi_1}/\sqrt{\left(\frac{\partial \Gamma(\xi)}{\xi_1}\right)^2 + \left(\frac{\partial \Gamma(\xi)}{\xi_2}\right)^2} & 0 \\ 0 & 0 & 1 \end{bmatrix} \quad (29)$$

Or

$$\tilde{E}^{3\times 3}(\xi) = \begin{bmatrix} \frac{\partial \Gamma(\xi)}{\xi_1}/\sqrt{\left(\frac{\partial \Gamma(\xi)}{\xi_1}\right)^2 + \left(\frac{\partial \Gamma(\xi)}{\xi_3}\right)^2} & \frac{\partial \Gamma(\xi)}{\xi_3}/\sqrt{\left(\frac{\partial \Gamma(\xi)}{\xi_1}\right)^2 + \left(\frac{\partial \Gamma(\xi)}{\xi_3}\right)^2} & 0 \\ 0 & 0 & 1 \\ \frac{\partial \Gamma(\xi)}{\xi_3}/\sqrt{\left(\frac{\partial \Gamma(\xi)}{\xi_1}\right)^2 + \left(\frac{\partial \Gamma(\xi)}{\xi_3}\right)^2} & -\frac{\partial \Gamma(\xi)}{\xi_1}/\sqrt{\left(\frac{\partial \Gamma(\xi)}{\xi_1}\right)^2 + \left(\frac{\partial \Gamma(\xi)}{\xi_3}\right)^2} & 0 \end{bmatrix} \quad (30)$$

or

$$\tilde{E}^{3\times3}(\xi) = \begin{bmatrix} 0 & 0 & 1 \\ \frac{\partial\Gamma(\xi)}{\xi_2}/\sqrt{\left(\frac{\partial\Gamma(\xi)}{\xi_2}\right)^2+\left(\frac{\partial\Gamma(\xi)}{\xi_3}\right)^2} & \frac{\partial\Gamma(\xi)}{\xi_3}/\sqrt{\left(\frac{\partial\Gamma(\xi)}{\xi_2}\right)^2+\left(\frac{\partial\Gamma(\xi)}{\xi_3}\right)^2} & 0 \\ \frac{\partial\Gamma(\xi)}{\xi_3}/\sqrt{\left(\frac{\partial\Gamma(\xi)}{\xi_2}\right)^2+\left(\frac{\partial\Gamma(\xi)}{\xi_3}\right)^2} & -\frac{\partial\Gamma(\xi)}{\xi_2}/\sqrt{\left(\frac{\partial\Gamma(\xi)}{\xi_2}\right)^2+\left(\frac{\partial\Gamma(\xi)}{\xi_3}\right)^2} & 0 \end{bmatrix} \quad (31)$$

For general consideration, we choose Eq.(29) as the selected pre-step orthogonal matrix. Thus the normal vector of $\tilde{E}^{3\times3}(\xi)$ is:

$$\tilde{e}_1(\xi) = \frac{1}{\sqrt{\left(\frac{\partial\Gamma(\xi)}{\xi_1}\right)^2+\left(\frac{\partial\Gamma(\xi)}{\xi_2}\right)^2}}\left[\frac{\partial\Gamma(\xi)}{\xi_1},\frac{\partial\Gamma(\xi)}{\xi_2},0\right]^T. \quad (32)$$

However, the desired normal vector of point outside the obstacle is:

$$e_1(\xi) = \frac{1}{\sqrt{\left(\frac{\partial\Gamma(\xi)}{\xi_1}\right)^2+\left(\frac{\partial\Gamma(\xi)}{\xi_2}\right)^2+\left(\frac{\partial\Gamma(\xi)}{\xi_3}\right)^2}}\left[\frac{\partial\Gamma(\xi)}{\xi_1},\frac{\partial\Gamma(\xi)}{\xi_2},\frac{\partial\Gamma(\xi)}{\xi_2}\right]^T. \quad (33)$$

To obtain $e_1(\xi)$ from $\tilde{e}_1(\xi)$, three rotation matrices $\tilde{R}_3$, $\tilde{R}_2$, $\tilde{R}_{-3}$ can be used. As seen in Fig. 4, $\tilde{R}_3$, $\tilde{R}_2$, $\tilde{R}_{-3}$ rotates $\tilde{e}_1(\xi)$ around $\xi_3$-axis, $\xi_2$-axis, $\xi_3$-axis, respectively. Then we have:

$$e_1(\xi) = \tilde{R}_{-3}\tilde{R}_2\tilde{R}_3\tilde{e}_1(\xi). \quad (34)$$

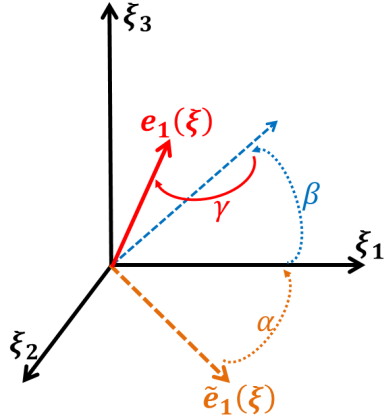

Fig. 4. Obtain $e_1(\xi)$ from $\tilde{e}_1(\xi)$ by rotating.

Then we can construct the orthogonal matrix $E^{3\times3}(\xi)$ by:

$$E^{3\times3}(\xi) = \tilde{R}_{-3}\tilde{R}_2\tilde{R}_3\tilde{E}^{3\times3}(\xi) = \begin{bmatrix} \frac{\partial\Gamma(\xi)}{\xi_1}\epsilon_3 & \frac{\partial\Gamma(\xi)}{\xi_2}\epsilon_2 & -\frac{\partial\Gamma(\xi)}{\xi_1}\frac{\partial\Gamma(\xi)}{\xi_3}\epsilon_2\epsilon_3 \\ \frac{\partial\Gamma(\xi)}{\xi_2}\epsilon_3 & -\frac{\partial\Gamma(\xi)}{\xi_1}\epsilon_2 & -\frac{\partial\Gamma(\xi)}{\xi_2}\frac{\partial\Gamma(\xi)}{\xi_3}\epsilon_2\epsilon_3 \\ \frac{\partial\Gamma(\xi)}{\xi_3}\epsilon_3 & 0 & \frac{\epsilon_3}{\epsilon_2} \end{bmatrix}. \quad (35)$$

where $\epsilon_3 \triangleq 1/\sqrt{\left(\frac{\partial\Gamma(\xi)}{\xi_1}\right)^2+\left(\frac{\partial\Gamma(\xi)}{\xi_2}\right)^2+\left(\frac{\partial\Gamma(\xi)}{\xi_3}\right)^2}$, $\epsilon_2 \triangleq 1/\sqrt{\left(\frac{\partial\Gamma(\xi)}{\xi_1}\right)^2+\left(\frac{\partial\Gamma(\xi)}{\xi_2}\right)^2}$,

$$\tilde{R}_3 = \begin{bmatrix} \frac{\partial \Gamma(\xi)}{\xi_1}\epsilon_2 & \frac{\partial \Gamma(\xi)}{\xi_2}\epsilon_2 & 0 \\ -\frac{\partial \Gamma(\xi)}{\xi_2}\epsilon_2 & \frac{\partial \Gamma(\xi)}{\xi_1}\epsilon_2 & 0 \\ 0 & 0 & 1 \end{bmatrix}. \tag{36}$$

$$\tilde{R}_2 = \begin{bmatrix} \frac{\epsilon_3}{\epsilon_2} & 0 & -\frac{\partial \Gamma(\xi)}{\xi_3}\epsilon_3 \\ 0 & 1 & 0 \\ \frac{\partial \Gamma(\xi)}{\xi_3}\epsilon_3 & 0 & \frac{\epsilon_3}{\epsilon_2} \end{bmatrix}. \tag{37}$$

$$\tilde{R}_{-3} = \begin{bmatrix} \frac{\partial \Gamma(\xi)}{\xi_1}\epsilon_2 & -\frac{\partial \Gamma(\xi)}{\xi_2}\epsilon_2 & 0 \\ \frac{\partial \Gamma(\xi)}{\xi_2}\epsilon_2 & \frac{\partial \Gamma(\xi)}{\xi_1}\epsilon_2 & 0 \\ 0 & 0 & 1 \end{bmatrix}. \tag{38}$$

**Proposition 2** $E^{3\times 3}(\xi)$ defined by Eq.(35) is a orthogonal matrix, i.e. $E^{3\times 3}(\xi) = E^{3\times 3}(\xi)^T$.

Due to one can rotate $E^{3\times 3}(\xi)$ around $e_1(\xi)$ axis, orthogonal basis matrix is not unique. For higher dimension, $E^{d\times d}(\xi)$ can be construct by $E^{(d-1)\times(d-1)}(\xi)$. The proposed modulation matrix can be defined by:

$$M^{d\times d}(\xi) = E^{d\times d}(\xi)D^{d\times d}(\xi)E^{d\times d}(\xi)^{(-1)} = E^{d\times d}(\xi)D^{d\times d}(\xi)E^{d\times d}(\xi)^T. \tag{39}$$

where $E^{d\times d}(\xi) \triangleq [e_1(\xi), e_2(\xi), \dots, e_d(\xi)]$ and $e_i(\xi)^T e_i(\xi) = 1, i = 1,2, \dots d$. The diagonal matrix $D^{d\times d}(\xi)$ is defined as

$$D^{d\times d}(\xi) = \begin{bmatrix} \lambda_1(\xi) & \cdots & 0 \\ \vdots & \ddots & \vdots \\ 0 & \cdots & \lambda_d(\xi) \end{bmatrix} = \begin{bmatrix} 1-\frac{1}{|\Gamma(\xi)|} & 0 & \cdots & 0 \\ 0 & 1+\frac{1}{|\Gamma(\xi)|} & \cdots & \cdots \\ \cdots & \cdots & \ddots & \cdots \\ 0 & \cdots & \cdots & 1+\frac{1}{|\Gamma(\xi)|} \end{bmatrix} \tag{40}$$

We can infer that:

$$\dot{\xi} = \|f(\xi)\|\left(1 - \frac{1}{|\Gamma(\xi)|}\right)\cos\left(\theta_{f(\xi),e_1(\xi)}\right)e_1(\xi) + \|f(\xi)\|\left(1 + \frac{1}{|\Gamma(\xi)|}\right)\sum_{i=2}^{d}\cos\left(\theta_{f(\xi),e_i(\xi)}\right)e_i(\xi)$$

$$\triangleq \sum_{i=1}^{d} C_i \cos\left(\theta_{f(\xi),e_i(\xi)}\right)e_i(\xi). \tag{41}$$

where $C_1 \geq 0, C_i > 0, i = 2,3,\dots d$. The radial component $C_1 \cos\left(\theta_{f(\xi),e_1(\xi)}\right)e_1(\xi) = 0$ when $C_1 = 0$, i.e., $\xi$ is on the surface of the obstacle. It ensures the motion never penetrate into the obstacle. From Eq.(41) we find that $\dot{\xi}$ is decided by angles between $f(\xi)$ and each vector of basis, i.e., $e_i(\xi), i = 1,2,\dots d$. If $\theta_{f(\xi),e_i(\xi)} > \pi/2$, then the basis vector $e_i(\xi)$ will lead $\dot{\xi}$ be $-e_i(\xi)$ at the $i$-th axis, and vice versa. If $\theta_{f(\xi),e_i(\xi)} = \pi/2$, then the basis vector $e_i(\xi)$ has no influence on $\dot{\xi}$ at the $i$-th axis.

*I. Another Modulation matrix formation*

From the above, we can infer:

$$M^{d\times d}(\xi) = E^{d\times d}(\xi)D^{d\times d}(\xi)E^{d\times d}(\xi)^T = I - \frac{e_1(\xi)e_1(\xi)^T}{|\Gamma(\xi)|} + \sum_{i=2}^{d}\frac{e_i(\xi)e_i(\xi)^T}{|\Gamma(\xi)|} \quad (42)$$

Then,

$$\dot{\xi} = f(\xi) - \frac{\|f(\xi)\|}{|\Gamma(\xi)|}\cos(\theta_{f(\xi),e_1(\xi)})e_1(\xi) + \frac{\|f(\xi)\|}{|\Gamma(\xi)|}\sum_{i=2}^{d}\cos(\theta_{f(\xi),e_i(\xi)})e_i(\xi). \quad (43)$$

$\dot{\xi}$ can be obtained by the addition and subtraction of vectors. $e_1(\xi)$ based vector is used to ensure no motions ($\{\xi_t\}, t = 0 \dots \infty$) can penetrate into the obstacle, and $e_i(\xi), i = 2, \dots d$ based vectors are used to make the motion move depart from the obstacle.

### IV. MANIPULATING ORTHOGONAL COORDINATES

Eq.(41) shows that $\dot{\xi}$'s direction is only decided by the angles between $f(\xi)$ and the column vectors of $E^{d\times d}(\xi) = [e_1(\xi), e_2(\xi), \dots, e_d(\xi)]$. When angles between $f(\xi)$ and $e_1(\xi), e_2(\xi), \dots, e_d(\xi)$ equal $\pi, \frac{\pi}{2}, \dots, \frac{\pi}{2}$ respectively, the local minimal problem occurs. If local minimal problem occurs, we have $\dot{\xi} \to 0$ when $\Gamma(\xi) \to 1$. In this section, we introduce the method to solve the local minimal problem by manipulating $E^{d\times d}(\xi)$.

*A. Rotation matrices*

When $d = 2$, the rotation matrix is defined by:

$$R^{2\times 2} = \begin{bmatrix} \cos(\theta) & -\sin(\theta) \\ \sin(\theta) & \cos(\theta) \end{bmatrix}. \quad (44)$$

To avoid the local minimal problem, we can just rotate the $E^{2\times 2}(\xi)$ by $\theta$ (rotate anticlockwise when $\theta > 0$, and vice versa). $\theta$ is defined by:

$$\theta = Y\delta_1\theta_{f(\xi),e_1(\xi)}(1 - \frac{1}{|\Gamma(\xi)|^{1/\delta_2}}). \quad (45)$$

where $\delta_1 \leq 1, \delta_2 \geq 1$ are the tuning parameters, $0 \leq \theta_{f(\xi),e_1(\xi)} \leq \pi$ is the angle between $f(\xi)$ and $e_1(\xi)$, $Y \in \{1, -1\}$ is indicator function. For the case of Fig. 2, the motion's trajectory will avoid the obstacle from top of the obstacle when $Y = -1$, while avoid the obstacle from bottom of the obstacle when $Y = 1$. Obstacle impenetrability still holds because of $\theta \to 0$ when $\Gamma(\xi) \to 1$.

When $d = 3$, we can obtain rotation matrix from arbitrary axis and angle [17] by:

$R^{3\times 3}(e_i(\xi), \theta_i) =$

$$\begin{bmatrix} e_{i1}^2(\xi)(1-\varphi_i^c)+\varphi_i^c & e_{i1}(\xi)e_{i2}(\xi)(1-\varphi_i^c)+e_{i3}(\xi)\varphi_i^s & e_{i1}(\xi)e_{i3}(\xi)(1-\varphi_i^c)-e_{i2}(\xi)\varphi_i^s \\ e_{i1}(\xi)e_{i2}(\xi)(1-\varphi_i^c)-e_{i3}(\xi)\varphi_i^s & e_{i2}^2(\xi)(1-\varphi_i^c)+\varphi_i^c & e_{i2}(\xi)e_{i3}(\xi)(1-\varphi_i^c)+e_{i1}(\xi)\varphi_i^s \\ e_{i1}(\xi)e_{i3}(\xi)(1-\varphi_i^c)+e_{i2}(\xi)\varphi_i^s & e_{i2}(\xi)e_{i3}(\xi)(1-\varphi_i^c)-e_{i1}(\xi)\varphi_i^s & e_{i3}^2(\xi)(1-\varphi_i^c)+\varphi_i^c \end{bmatrix}$$

$, i = 1,2,3 \quad (46)$

where $\cos(\theta_i) \triangleq \varphi_i^c, \sin(\theta_i) \triangleq \varphi_i^s, e_i(\xi) \triangleq [e_{i1}(\xi), e_{i2}(\xi), e_{i3}(\xi)]^T$.

As seen in Fig. 5, $R^{3\times 3}(e_i(\xi), \theta_i)$ works by rotating $\theta_i$ around the $e_i(\xi)$ axis, where $\theta_i$ is

defined by:

$$\theta_i = Y\delta_{1i}\theta_{f(\xi),e_1(\xi)}\left(1 - \frac{1}{|\Gamma(\xi)|^{1/\delta_{2i}}}\right), i = 2,3. \tag{47}$$

where $\delta_{1i} \leq 1, \delta_{2i} \geq 1$ are the tuning parameters, $0 \leq \theta_{f(\xi),e_1(\xi)} \leq \pi$ is the angle between $f(\xi)$ and $e_1(\xi)$.

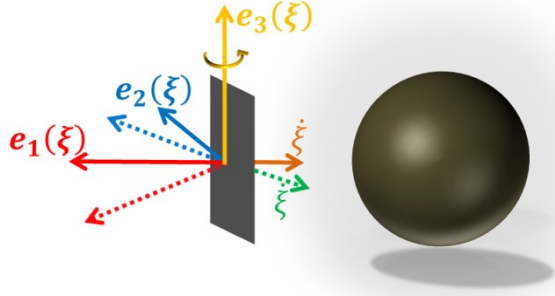

Fig. 5. Example for solving the local minimal problem by rotating $E^{3\times 3}$ around $e_3(\xi)$.

The rotation matrix around $e_i(\xi)$ can also be denoted as $R^{3\times 3}(e_j(\xi), e_k(\xi), \theta_{jk}), j \neq k, k \neq i, i \neq j$, which means rotate the $e_j(\xi) - O - e_k(\xi)$ plane. When $d > 3$, one can use Givens rotation [18] to conduct the rotation matrices to rotate planes.

In summary, the rotation matrix can be denote as:

$$R^{d\times d}(\xi) = \prod_{j,k=2,\ldots,d, j\neq k} R^{d\times d}\left(e_j(\xi), e_k(\xi), \theta_{jk}\right). \tag{48}$$

Then the orthogonal basis matrix can be remedy by:

$$\bar{E}^{d\times d}(\xi) = R^{d\times d}(\xi)E^{d\times d}(\xi). \tag{49}$$

Thus, we have

$$M^{d\times d}(\xi) = \bar{E}^{d\times d}(\xi)D^{d\times d}(\xi)\bar{E}^{d\times d}(\xi)^{(-1)} = \bar{E}^{d\times d}(\xi)D^{d\times d}(\xi)\bar{E}^{d\times d}(\xi)^T. \tag{50}$$

**Proposition 3** Any trajectory $\{\xi_t\}, t = 0, \ldots, \infty$, that starts outside the obstacle according to (50), will never penetrate the obstacle.

B. *Combine different coordinates*

For 3-D case, in certain situations, $\left(\frac{\partial \Gamma(\xi)}{\xi_1}\right)^2 + \left(\frac{\partial \Gamma(\xi)}{\xi_2}\right)^2 \neq 0$, $\left(\frac{\partial \Gamma(\xi)}{\xi_1}\right)^2 + \left(\frac{\partial \Gamma(\xi)}{\xi_3}\right)^2 \neq 0$ and $\left(\frac{\partial \Gamma(\xi)}{\xi_2}\right)^2 + \left(\frac{\partial \Gamma(\xi)}{\xi_3}\right)^2 \neq 0$ could not establish at the same time. Thus, we may attempt to use Eq.(36)~Eq.(38) as the pre-step orthogonal basis due to various requirements of dynamical system. Different pre-step orthogonal matrix lead to different orthogonal matrix, which can be denoted by $E^{3\times 3}_{<i>}(\xi), i = 1,2,3$. For example, $E^{3\times 3}_{<1>}(\xi)$ requires $\left(\frac{\partial \Gamma(\xi)}{\xi_2}\right)^2 + \left(\frac{\partial \Gamma(\xi)}{\xi_3}\right)^2 \neq 0$. The correspond matrices can be denoted by $R^{3\times 3}_{<i>}(\xi)$, $\bar{E}^{3\times 3}_{<i>}(\xi)$ in Eq.(48) and Eq.(49).

Then, the total modulation matrix can be defined by:

$$M^{3\times3}(\xi) = \sum_{i=1}^{3} \eta_i \bar{E}^{3\times3}_{<i>}(\xi) D^{d\times d}(\xi) \bar{E}^{3\times3}_{<i>}(\xi)^T. \tag{51}$$

where $\eta_i, i = 1,2,3$ are tuning parameters.

For higher dimensions, $M^{d\times d}(\xi)$ can be defined by multiple coordinates to enhance its robustness.

C. *Reactivity and tail effect*

In [1], reactivity is considered by modifying $D^{d\times d}(\xi) = diag[\lambda_1(\xi),..,\lambda_d(\xi)]$ as follows:

$$\begin{cases} \lambda_1(\xi) = 1 - \dfrac{1}{|\Gamma(\xi)|^{1/\rho}} \\ \lambda_i(\xi) = 1 + \dfrac{1}{|\Gamma(\xi)|^{1/\rho}}, 2 \leq i \leq d \end{cases}. \tag{52}$$

where $\rho > 0$ is the reactivity parameter. We usually set $\rho = 1$ in this paper for we focus on the rotation matrices' ability to respond to the presence of an obstacle.

Tail effect is reduced in [1] by remedy $\lambda_1(\xi) = 1$ when the trajectory is going away from the obstacle, i.e., $e_1(\xi)^T f(\xi) \geq 0$. In the view of the proposed approach, the tail effect is reduced by enhance the influence of basis $e_1(\xi)$ $(1 > 1 - \frac{1}{|\Gamma(\xi)|})$ when $\theta_{f(\xi),e_1(\xi)} \leq \pi/2$.

D. *Consistency of $f(\xi_t)$ and $\dot{\xi}_t$*

To improve the consistency of the motions $(\{\xi_t\}, t = 0 ... \infty)$, we consider the consistency of $f(\xi_t)$ and $\dot{\xi}_t (t = 0 ... \infty)$. At time $T$, $\dot{\xi}_T$ is computed to generate $\xi_{T+1}$ from $\xi_T$. However $f(\xi_{T+1})$'s direction has no relationship with $\dot{\xi}_T$ and $M_T$. At time $T+1$, suppose we have one obstacle, we can modify $D^{d\times d}(\xi) = diag[\lambda_1(\xi),..,\lambda_d(\xi)]$ by

$$\begin{cases} \lambda_1(\xi) = 1 - \dfrac{\mu_j}{|\Gamma(\xi)|} \\ \lambda_i(\xi) = 1 + \dfrac{\mu_j}{|\Gamma(\xi)|}, 2 \leq i \leq d \end{cases}, j = 0,1,...,T+1. \tag{53}$$

where $\mu_j, j = 0,1,...T+1$ is the weighting parameter of the modulation matrix at time $j$. $\mu_{T+1} = 1$ and $\mu_{T+1} \geq \mu_T \geq \cdots \geq \mu_0$.

Then we use the modulation matrices from time 0 to time $T$ to estimate the new length and the direction of $f(\xi_{T+1})$. That is to say, all time modulation matrix are utilized to estimate new $f(\xi_{T+1})$ (denoted as $f_{new}(\xi_{T+1})$) by:

$$f_{new}(\xi_{T+1}) = (\prod_{i=0}^{T} M_i) f(\xi_{T+1}). \tag{54}$$

Then,

$$\dot{\xi}_{T+1} = M_{T+1} f_{new}(\xi_{T+1}) = (\prod_{i=0}^{T+1} M_i) f(\xi_{T+1}). \tag{55}$$

We have:

$$\|f_{new}(\xi_{T+1}) - \dot{\xi}_T\| = \left\|\prod_{i=0}^{T} M_i\right\| \|f(\xi_{T+1}) - f(\xi_T)\|. \tag{56}$$

Apparently, $f_{new}(\xi_{T+1}) \to \dot{\xi}_T$ when $f(\xi_{T+1}) \to f(\xi_T)$, which ensures the consistency of $f(\xi_t)$ and $\dot{\xi}_t (t = 0 \ldots \infty)$.

If we adopt $D^{d \times d}(\xi)$ as:

$$\begin{cases} \lambda_1(\xi) = 1 - \dfrac{1}{|\Gamma(\xi)|} \\ \lambda_i(\xi) = 1 + \dfrac{1}{|\Gamma(\xi)|}, 2 \le i \le d \end{cases}, j = 0,1,\ldots,T+1. \tag{57}$$

Then, $\dot{\xi}_{T+1}$ can be defined as:

$$\dot{\xi}_{T+1} = \left(\sum_{j=1}^{T+1} \mu_j M_i\right) f(\xi_{T+1}). \tag{58}$$

Considering Eq.(53)~Eq.(55), if we set $\mu_{T+1} = \mu_T = 1$, $\mu_j = 0, j = 0,1,\ldots T-1$. We have

$$\|f_{new}(\xi_{T+1})\| = \|M_T\|\|f(\xi_{T+1})\| = (1 - \frac{1}{|\Gamma(\xi)|})(1 + \frac{1}{|\Gamma(\xi)|})^{d-1}\|f(\xi_{T+1})\|. \tag{59}$$

Obviously, $\|f_{new}(\xi_{T+1})\| \to 0$ when the motion move towards to the obstacle. Moreover, $\Theta(|\Gamma(\xi)|, d) \triangleq (1 - \frac{1}{|\Gamma(\xi)|})(1 + \frac{1}{|\Gamma(\xi)|})^{d-1}$ affects the magnitude of $f_{new}(\xi_{T+1})$, as seen in Fig. 6.

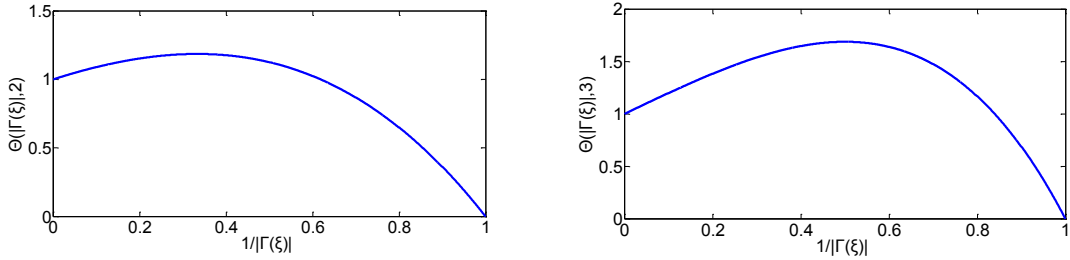

a. The value of $\Theta(|\Gamma(\xi)|, 2)$        b. The value of $\Theta(|\Gamma(\xi)|, 3)$

Fig. 6. $\Theta(|\Gamma(\xi)|, 2)$ and $\Theta(|\Gamma(\xi)|, 3)$.

*E. Trap Area*

The indicator function $Y \in \{1, -1\}$ In Eq.(45) effects the motion's trajectory for obstacle avoidance, i.e., the preference of the motion to go top around or under around the obstacle. If multiple convex obstacles have intersections, then the trap area may appear. To handle with this tough problem, we can define the indicator functions and weights of those connected obstacles to reduce the chance of the motion dropping into the Trap Area.

Suppose we have $N_c$ connected obstacles, we can compute $\Gamma_{(i)}(\xi), i = 1,2,\ldots N_c$. The weights of these obstacles are defined as:

$$w_j = \begin{cases} 1, j = \underset{i}{argmin}\, \Gamma_{(i)}(\xi), i = 1,2,\ldots N_c \\ 0, \quad otherwise \end{cases}. \tag{60}$$

## V. EXPERIMENT RESULTS

In this section, we evaluate the performance of the proposed dynamical system based Obstacle

Avoidance via Manipulating Orthogonal Coordinates (OA-MOC). All experiment tools are based on [1].

---
**Algorithm 1 Procedure and parameters of OA-MOC for all experiments**
---
1. Compute $E^{2\times 2}(\xi)$ by Eq.(28) or $E^{3\times 3}(\xi)$ by Eq.(35).
---
2. At time $T+1$, we compute $D_{(i)}{}^{d\times d}(\xi)$ by

$$\begin{cases} \lambda_1(\xi) = 1 - \dfrac{\mu_j w_i}{|\Gamma(\xi)|} \\ \lambda_i(\xi) = 1 + \dfrac{\mu_j w_i}{|\Gamma(\xi)|}, 2 \le i \le d \end{cases}, j = 0,1,\dots,T+1$$

where $w_i, i = 1,2,\dots,N$ is the $i$-th obstacle's weight. We set $\mu_{T+1} = \mu_T = 1$, $\mu_j = 0, j = 0,1,\dots,T-1$. If obstacles have no intersections, $w_i$ is computed by Eq(24). Weights of obstacles with intersections are computed by Eq.(60). The tail effect reduce strategy is always utilized in all experiments.
---
3. Compute $\theta$ by Eq.(45) when $d = 2$. Compute $\theta_i$ by Eq.(47) when $d = 3$. We set $\delta_1 = \delta_{12} = \delta_{13} = 1/2$, $\delta_2 = \delta_{22} = \delta_{23} = 2$.
---
4. Compute $R^{d\times d}$ by Eq.(54)(44), Eq.(46), Eq.(48).
---
5. Compute $\bar{E}^{d\times d}(\xi)$ by Eq.(49).
---
6. Compute $M_{T+1}{}^{d\times d}(\xi)$ by Eq.(54) (50)
---
7. Compute $f(\xi_{T+1})$ by Eq.(54)
---
8. Compute $\dot{\xi}_{T+1}$ by $\dot{\xi}_{T+1} = M_{T+1}{}^{d\times d}(\xi) f(\xi_{T+1})$.
---

### A. Local minimal problem

A simple autonomous linear system is defined by: $f(\xi) = -(\xi - \xi_{goal})$, where $\xi_{goal} = [0,0]^T$. The obstacle is defined by $\Gamma(\xi) = (\frac{\xi_1+9}{3.6})^2 + (\frac{\xi_2}{3.6})^2$. As seen in Fig. 7(a), the DS starts from $[-18,0]^T$. Then, the local minimal problem occurs, as seen in Fig. 7(b). [1] proposed a contouring approach to deal with this problem, as seen in Fig. 7(c). The proposed OA-MOC approach's performance is illustrated in Fig. 7(d)-(f), which depicts the local minimal problem is solved.

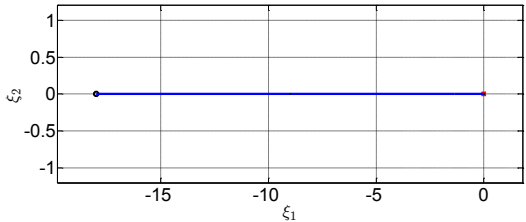

(a) start from $[-18,0]^T$ to $[0,0]^T$

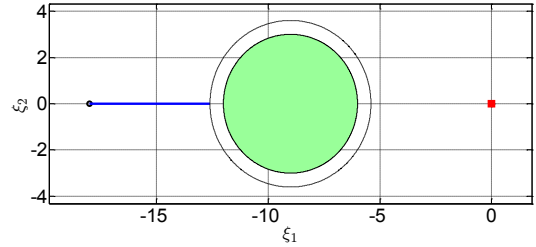

(b) Local minimal problem by [1]

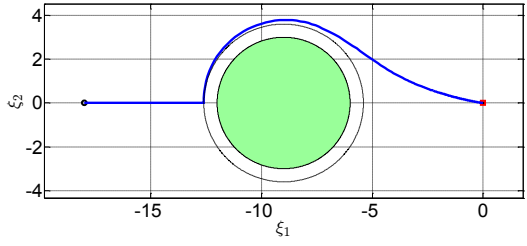
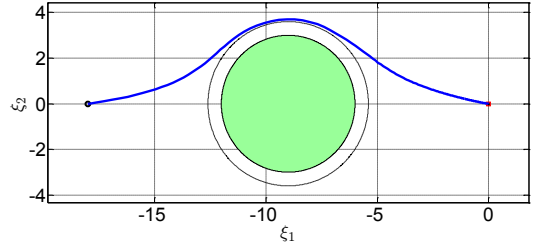

(c)Contouring method in [1]    (d)The proposed method without motion consistency

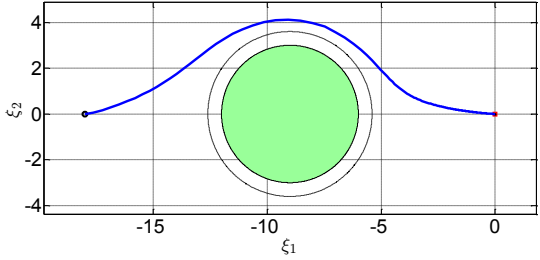
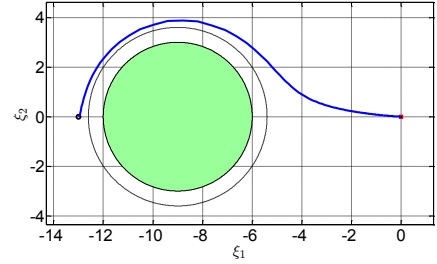

(e) The proposed method with motion consistency    (f) The proposed method start from $[-13,0]^T$

Fig. 7. Solve the local minimal problem.

When the obstacle is defined by $\Gamma(\xi) = (\frac{\xi_1+9}{3.6})^4 + (\frac{\xi_2}{3.6})^4$, the performance of [1] and the proposed approach are illustrated in Fig. 8.

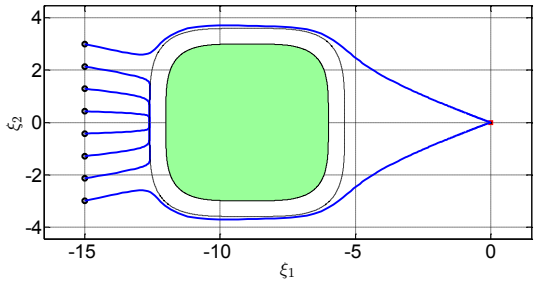
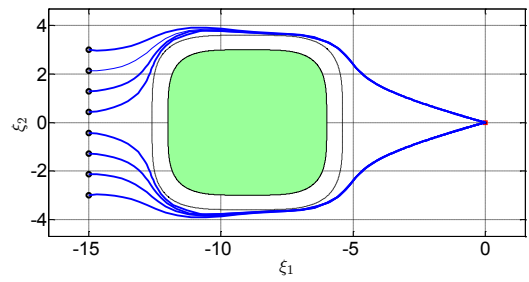

(a)Method in [1]    (b) OA-MOC. $\Upsilon = -1$ when $\xi_2 > 0$ and $\Upsilon = 1$ when $\xi_2 < 0$.

Fig. 8. Performance of obstacle avoidance by two approaches.

*B. One obstacle*

The obstacles are defined by $\Gamma(\xi) = \left(\frac{\xi_1+9}{3.6}\right)^2 + \left(\frac{\xi_2}{3.6}\right)^2$. The performance of [1] and the proposed approach are illustrated in Fig. 9.

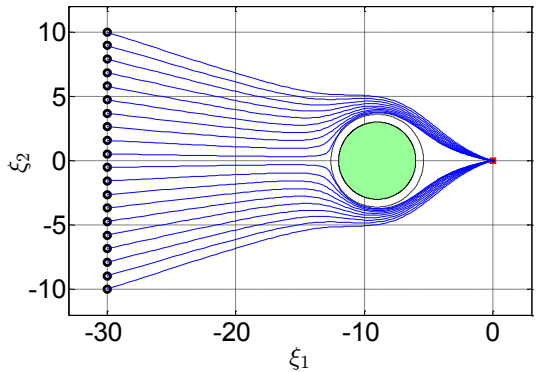
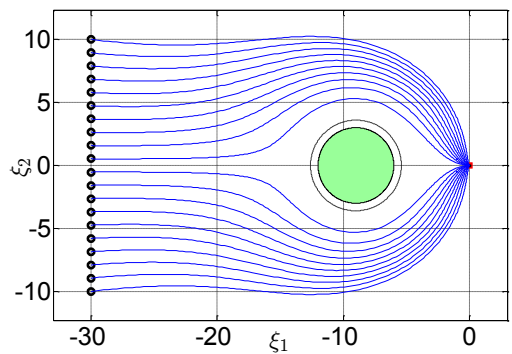

(a)Method in [1] with $\rho = 1$    (b)Method in [1] with $\rho = 5$

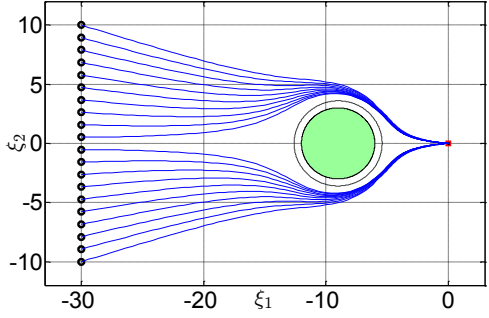 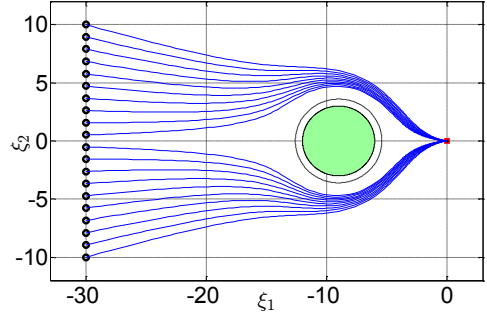

(c) OA-MOC with $\delta_2 = 2$. $Y = -1$ when $\xi_2 > 0$ and $Y = 1$ when $\xi_2 < 0$.

(d) OA-MOC with $\delta_2 = 8$.

Fig. 9. Performance of one obstacle avoidance by two approaches.

### C. Multiple obstacles

The obstacles are defined by $\Gamma_{(1)}(\xi) = \left(\frac{\xi_1+5}{3.6}\right)^2 + \left(\frac{\xi_2}{3.6}\right)^2$, $\Gamma_{(2)}(\xi) = \left(\frac{\xi_1+12}{3.6}\right)^2 + \left(\frac{\xi_2-3}{3.6}\right)^2$, $\Gamma_{(3)}(\xi) = (\frac{\xi_1+15}{3.6})^2 + (\frac{\xi_2+5}{3.6})^2$. The performance of [1] and the proposed approach are illustrated in Fig. 10. For the $i$-th obstacle, $Y_{(i)} = -1$ if $\xi$ is beyond the line that construct by the goal point and the center point of the $i$-th obstacle, $Y_{(i)} = 1$ otherwise.

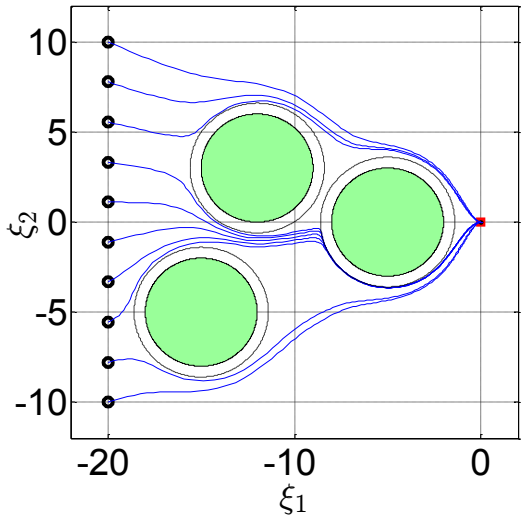 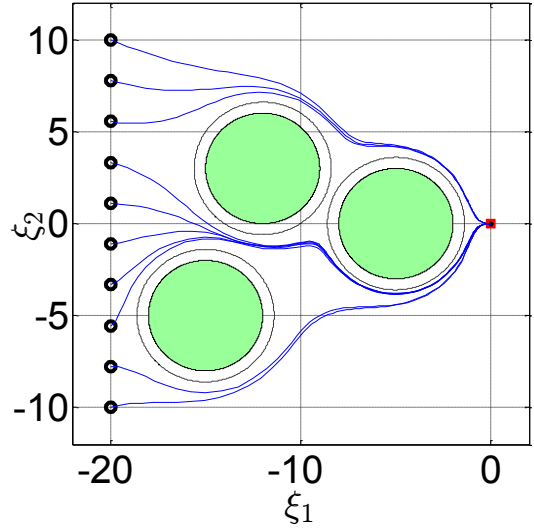

(a) Method in [1]

(b) OA-MOC.

Fig. 10. Performance of multiple obstacles avoidance by two approaches.

### D. Trap area problem

The obstacles are defined by $\Gamma_{(1)}(\xi) = \left(\frac{\xi_1+9}{3.6}\right)^2 + \left(\frac{\xi_2-3}{3.6}\right)^2$, $\Gamma_{(2)}(\xi) = \left(\frac{\xi_1+9}{3.6}\right)^2 + \left(\frac{\xi_2-1}{3.6}\right)^2$, $\Gamma_{(3)}(\xi) = (\frac{\xi_1+9}{3.6})^2 + (\frac{\xi_2+3}{3.6})^2$, $\Gamma_{(4)}(\xi) = (\frac{\xi_1+9}{3.6})^2 + (\frac{\xi_2+1}{3.6})^2$. The performance of [1] and the proposed approach are illustrated in Fig. 11, Trap Areas exist in this case.

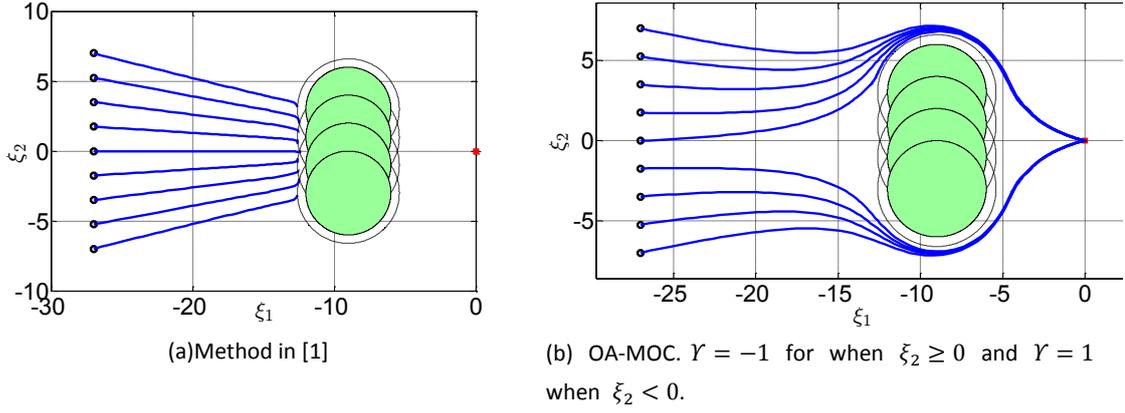

(a) Method in [1]

(b) OA-MOC. $Y = -1$ for when $\xi_2 \geq 0$ and $Y = 1$ when $\xi_2 < 0$.

Fig. 11. Performance of obstacle avoidance by two approaches where Trap Areas exist.

The performance of OA-MOC to deal with Trap Area is also illustrated in Fig. 12. Motions start from yellow marked points fail avoiding obstacles due to their positions.

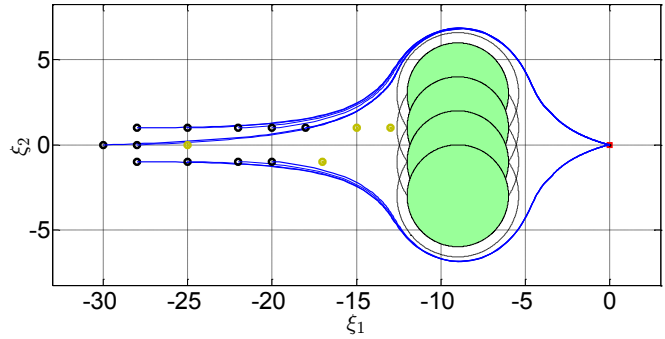

Fig. 12. The performance of OA-MOC to deal with Trap Area

*E. 3-D case*

We test the effect of the proposed rotation matrix in 3-D by a simple case. The obstacles are defined by $\Gamma(\xi) = \left(\frac{\xi_1+9}{3.6}\right)^2 + \left(\frac{\xi_2}{3.6}\right)^2 + \left(\frac{\xi_2}{3.6}\right)^3$. The performance of [1] and the proposed approach are illustrated in Fig. 13. Fig. 13.(a) shows the performance of [1]. Fig. 13.(b) shows the performance of OA-MOC that rotating coordinates only around the $e_3(\xi)$-axis by $R^{3\times3}(\xi) = R^{d\times d}(e_1(\xi), e_2(\xi), \theta_{12})$. Fig. 13.(c) shows the performance of OA-MOC that rotating coordinates only around the $e_2(\xi)$-axis by $R^{3\times3}(\xi) = R^{d\times d}(e_1(\xi), e_3(\xi), \theta_{13})$. Fig. 13.(d) shows the performance of OA-MOC that rotating coordinates around the $e_2(\xi)$-axis and the $e_3(\xi)$-axis by $R^{3\times3}(\xi) = R^{d\times d}(e_1(\xi), e_3(\xi), \theta_{13})R^{d\times d}(e_1(\xi), e_2(\xi), \theta_{12})$.

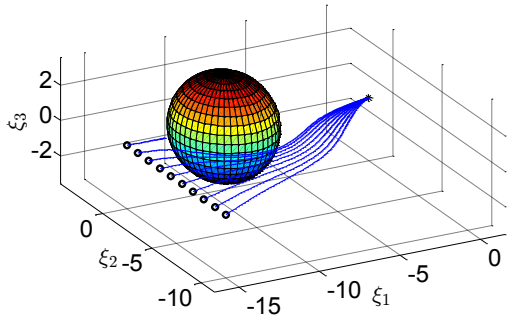

(a) Method in [1]

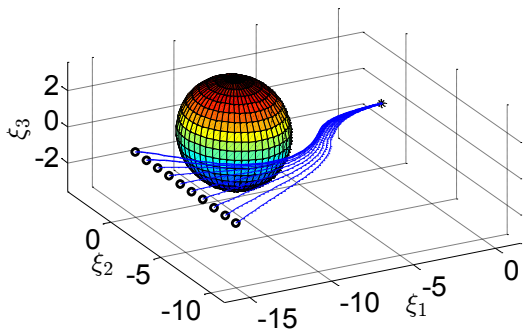

(b) OA-MOC, coordinates rotate only around the

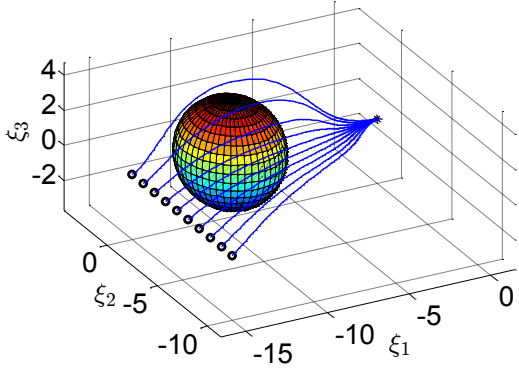
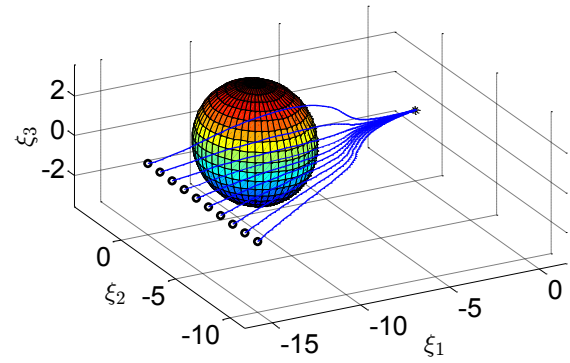

(c) OA-MOC, coordinates rotate only around the $e_2(\xi)$-axis, $Y = -1$

(d) OA-MOC, coordinates rotate around $e_2(\xi)$-axis and the $e_3(\xi)$-axis, $Y = -1$

Fig. 13. A 3-D case of obstacle avoidance.

### F. A case of cycle dynamic system

A simple cycle dynamic system is defined by: $f(\xi_1) = \xi_2 - \xi_1(\xi_1^2 + \xi_2 sin\xi_1 - 1)$, $f(\xi_2) = -\xi_1 - \xi_2(\xi_1^2 + \xi_2 sin\xi_1 - 1)$, $f(\xi_3) = 0$, and $f(\xi) = 10f(\xi)/\|f(\xi)\|$. Obstacles are defined as $\Gamma_{(1)}(\xi) = \left(\frac{\xi_1}{3.6}\right)^2 + \left(\frac{\xi_2}{3.6}\right)^2 + \left(\frac{\xi_3}{3.6}\right)^2$, $\Gamma_{(2)}(\xi) = \left(\frac{\xi_1}{3.6}\right)^2 + \left(\frac{\xi_2}{3.6}\right)^2 + \left(\frac{\xi_3-10}{3.6}\right)^2$, $\Gamma_{(3)}(\xi) = \left(\frac{\xi_1+10}{3.6}\right)^2 + \left(\frac{\xi_2}{3.6}\right)^2 + \left(\frac{\xi_3}{3.6}\right)^2$, and we start from $[-15,0,0]^T$, as seen in Fig. 14.

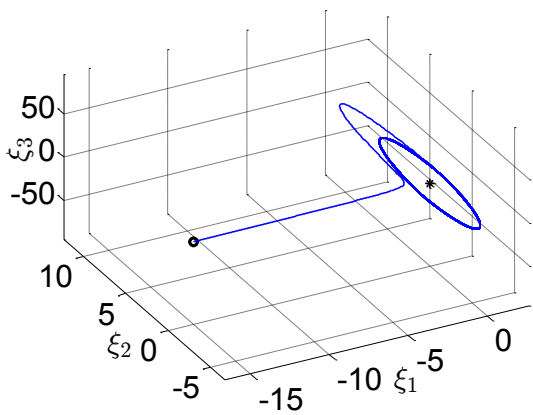
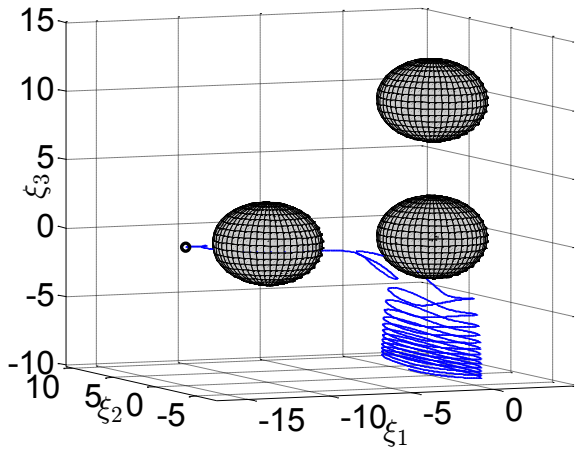

(a) cycle dynamic system without obstacles

(b) Method in [1]

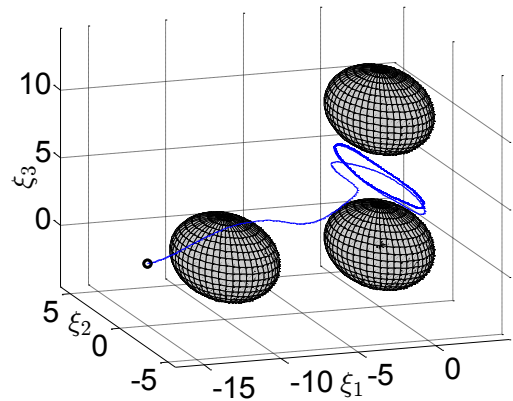
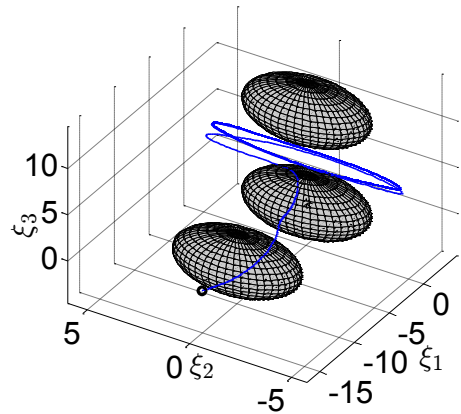

(c) OA-MOC                                    (d) Another view of OA-MOC

Fig. 14. A 3-D case obstacle avoidance of cycle dynamic system with 3000 iterations.

## G. Patrol ability

In fact, the proposed method can also be used for patrolling. In this application, the dynamic system is described as

$$\dot{\xi}_{t+1} = M^{d\times d}(\xi)\dot{\xi}_t, \qquad \dot{\xi}_{t+1} = \dot{\xi}_{t+1}/\|\dot{\xi}_{t+1}\|. \tag{61}$$

where $\dot{\xi}_0$ can be set as $\dot{\xi}_0 = \xi_{center} - \xi_0$.

The desire trajectory can be calculated by integrating $\dot{\xi}_t$:

$$\xi_{t+1} = \xi_t + \dot{\xi}_t \delta_t. \tag{62}$$

The proposed method for patrol is named as Patrol via Manipulating Orthogonal Coordinates (P-MOC). The patrol ability of the proposed method and method in [1] is shown in Fig. 15, where P-MOC shows the convergence of the patrolling trajectory for several convex shapes.

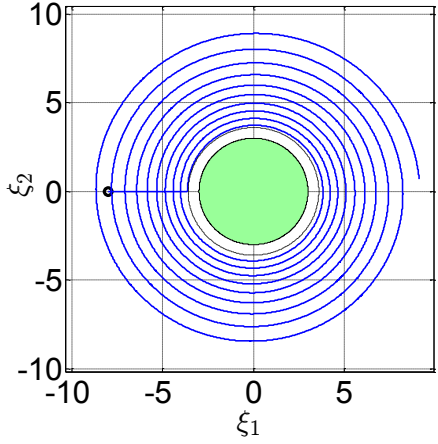

(a) Method in [1], where $\Gamma(\xi) = \left(\frac{\xi_1}{3.6}\right)^2 + \left(\frac{\xi_2}{3.6}\right)^2$.

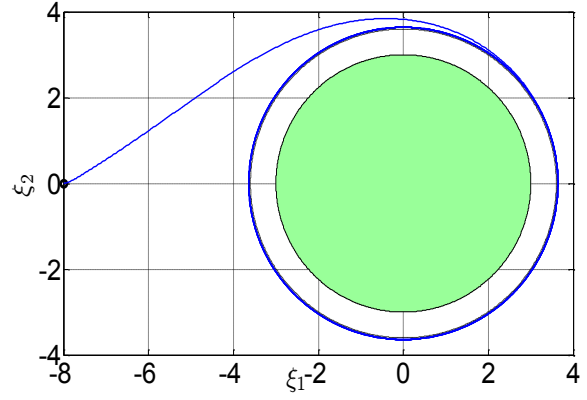

(b) The proposed P-MOC, where $\delta_1 = 1/2$, $\delta_2 = 1$, $\Gamma(\xi) = \left(\frac{\xi_1}{3.6}\right)^2 + \left(\frac{\xi_2}{3.6}\right)^2$.

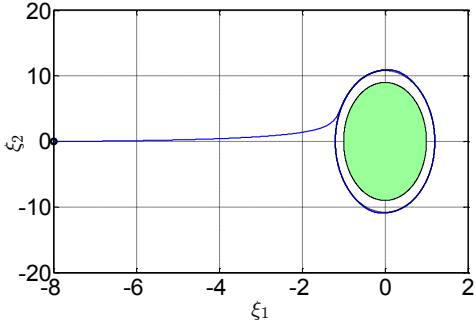

(c) The proposed P-MOC, where $\delta_1 = 1/2$, $\delta_2 = 1$, $\Gamma(\xi) = \left(\frac{\xi_1}{1.2}\right)^2 + \left(\frac{\xi_2}{10.8}\right)^2$.

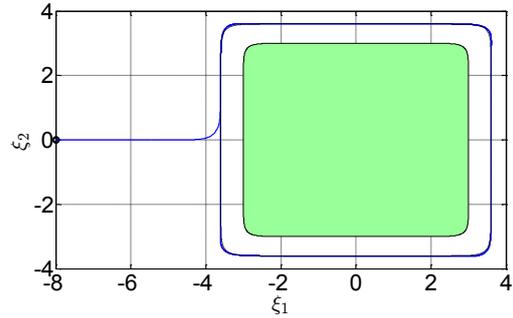

(d) The proposed P-MOC, where $\delta_1 = 1/2$, $\delta_2 = 1$, $\Gamma(\xi) = \left(\frac{\xi_1}{3.6}\right)^{16} + \left(\frac{\xi_2}{3.6}\right)^{16}$.

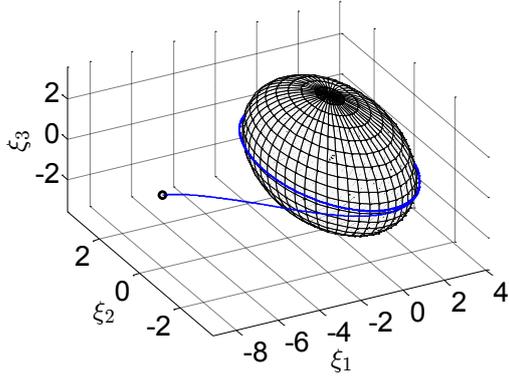 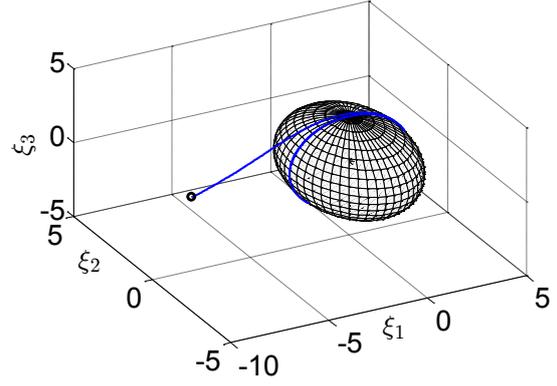

(e) Rotate in horizon plane by the proposed P-MOC, $\Gamma(\xi) = \left(\frac{\xi_1}{3}\right)^2 + \left(\frac{\xi_2}{3}\right)^2 + \left(\frac{\xi_3}{3}\right)^2$. According to Eq.(51), $\bar{E}_{<3>}^{3\times3}(\xi)$ is adopt.

(f) Rotate in vertcial plane by the proposed P-MOC, $\Gamma(\xi) = \left(\frac{\xi_1}{3}\right)^2 + \left(\frac{\xi_2}{3}\right)^2 + \left(\frac{\xi_3}{3}\right)^2$. According to Eq.(51), $\bar{E}_{<2>}^{3\times3}(\xi)$ is adopt.

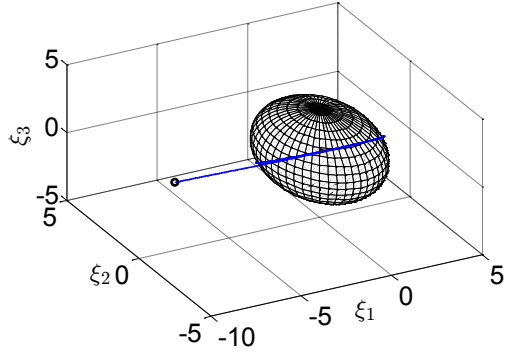 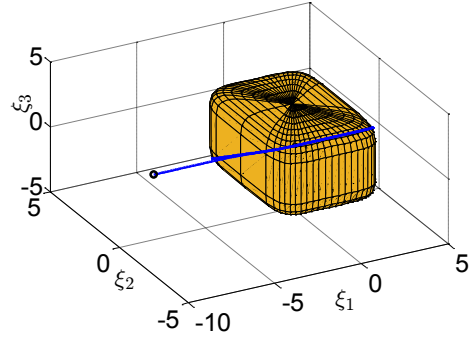

(e) Rotate in oblique plane by the proposed P-MOC, $\Gamma(\xi) = \left(\frac{\xi_1}{3}\right)^2 + \left(\frac{\xi_2}{3}\right)^2 + \left(\frac{\xi_3}{3}\right)^2$. According to Eq.(51), $\bar{E}_{<2>}^{3\times3}(\xi)$ and $\bar{E}_{<3>}^{3\times3}(\xi)$ are adopt, where $\eta_2 = \eta_3 = 1$.

(e) Rotate in oblique plane by the proposed P-MOC, $\Gamma(\xi) = \left(\frac{\xi_1}{3}\right)^6 + \left(\frac{\xi_2}{3}\right)^6 + \left(\frac{\xi_3}{3}\right)^6$. According to Eq.(51), $\bar{E}_{<2>}^{3\times3}(\xi)$ and $\bar{E}_{<3>}^{3\times3}(\xi)$ are adopt, where $\eta_2 = \eta_3 = 1$.

Fig. 15. Patrolling experiments.

## VI. CONCLUSION AND FUTURE WORK

In this paper, we proposed a novel method for obstacle avoidance based on dynamical system. Firstly, we studied the basic form of the modulation matrix, which is more comprehensive by using orthogonal coordinates. The new direction of the motion can be expressed as a linear combination of a set of basis vectors. Secondly, we manipulated these basis vectors, which named orthogonal coordinates, by a proposed rotation function. Thirdly, we evaluated the proposed method on several basic cases, including 2-D or 3-D obstacle avoidance, solving local minimal problem, dealing with TA problem, patrolling, etc. More applications will to be investigated in our future work.